\documentclass{article}
\usepackage{spconf,amsmath,graphicx,hyperref}
\usepackage{cite}
\usepackage{amsmath,amssymb,amsfonts}
\usepackage{algorithmic}
\usepackage{graphicx}
\usepackage{textcomp}
\usepackage{xcolor}
\usepackage[ruled,linesnumbered]{algorithm2e}
\usepackage{multirow}
\usepackage{multicol}
\usepackage[table]{xcolor}
\usepackage{subfigure} 
\usepackage{booktabs}

\title{HMVI: Unifying Heterogeneous Attributes with Natural Neighbors for Missing Value Inference}
\name{Xiaopeng Luo$^{1}$, Zexi Tan$^{1}$, Zhuowei Wang$^{1,\star}$\thanks{$^{\star}$Corresponding author: Zhuowei Wang}}
  
  \address{$^{1}$ 
  School of Computer Science and Technology\\Guangdong University of Technology\\
Guangzhou, China}
\begin{document}
%
\maketitle
\begin{abstract}
Missing value imputation is a fundamental challenge in machine intelligence, heavily dependent on data completeness. Current imputation methods often handle numerical and categorical attributes independently, overlooking critical interdependencies among heterogeneous features. To address these limitations, we propose a novel imputation approach that explicitly models cross-type feature dependencies within a unified framework. Our method leverages both complete and incomplete instances to ensure accurate and consistent imputation in tabular data. Extensive experimental results demonstrate that the proposed approach achieves superior performance over existing techniques and significantly enhances downstream machine learning tasks, providing a robust solution for real-world systems with missing data.
\end{abstract}
\begin{keywords}
Missing values imputation, heterogeneous attributes, dissimilarity measure, natural neighbor.
\end{keywords}
\section{Introduction}
\label{sec:intro}
Missing values in real-world data (from sensor failures, transmission errors, etc.) challenge cluster analysis\cite{icassp2025}, as most conventional algorithms need complete datasets. Modern multimodal signal-sourced datasets, with complex nonlinear relationships and heterogeneous attributes (nominal, ordinal, and numerical), common in audio, sensor, and biomedical signal processing, exacerbate this issue\cite{chang2025valve}. Specialized imputation methods are needed due to varying performance across data types and missingness patterns.
\begin{figure}[t]
		\centering
		\includegraphics[width=1\linewidth]{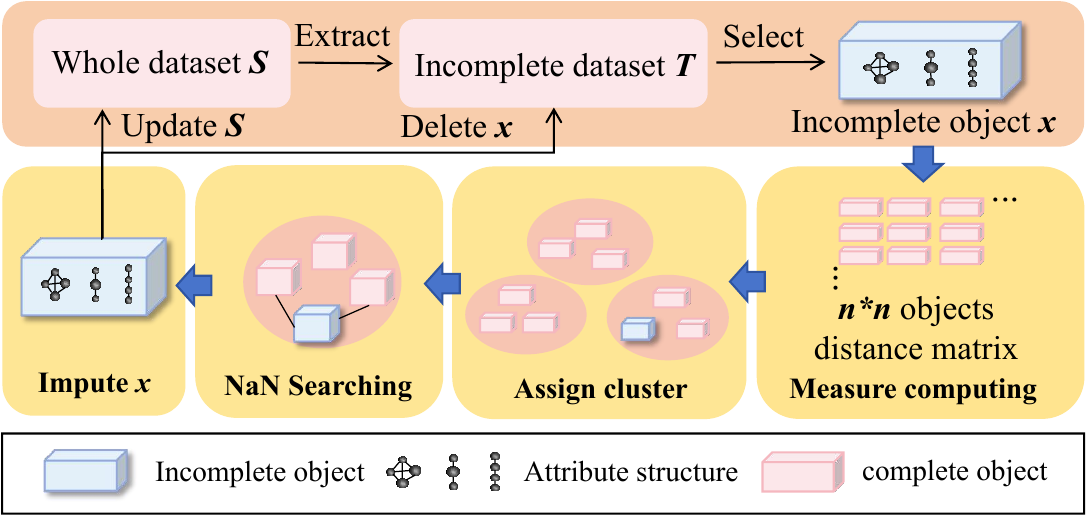}
		\caption{Mechanism of the proposed HMVI.}
		\label{fig:flow}
\end{figure}

Direct deletion \cite{deletion} (omitting incomplete data) and mean/ mode substitution \cite{ms} are simple but miss data and cause biases \cite{deletion2}. More popular approaches, like K-Nearest Neighbors-based Missing values Imputation (KNNMI)\cite{Knn,kntn} and K-Means Clustering-based Missing values Imputation(KMCMI) \cite{KMI} for numerical data, C4.5\cite{C4.5} and association rule\cite{Ar} for categorical data \cite{zhang2024cross,lan2025improve}. Imputation methods for heterogeneous data are rather rare in the literature. It was first introduced by Rubin \cite{firstmix} with multiple imputation. Then an approach based on maximum likelihood estimation is proposed, which combines the multivariate normal model for continuous and the Poisson/multinomial model for categorical data\cite{maxlikelihood,zhao2025break}. A more refined method, MICE\cite{MICE}, relies on tuning parameters or specifying a parametric model; missing values are repeatedly imputed to create a complete dataset. Another powerful MissForest\cite{forest} is based on Random forest(RF) \cite{RF}.

In general, most existing imputation methods are limited to one type of attribute. In addition, these methods only fill the missing values based on the observed complete samples, ignoring the information contained in the observed incomplete samples. When the dataset contains different types of attributes and a large number of incomplete samples, the effectiveness of these methods is significantly reduced.

This paper, therefore, proposes a new missing value inference method named Heterogeneous attributes Missing Value Inference (HMVI). We first study the connection between heterogeneous attributes and define a dissimilarity measure for the heterogeneous data with missing values. Then, we present a clustering method for incomplete heterogeneous datasets. Finally, we propose to infer the missing values according to the natural neighbor relationship in the clusters. Comparison of experimental results shows the excellent performance of the proposed method.

 The main contributions of this paper are three-fold:
	\begin{itemize}
		\item \textbf{Unified Heterogeneous Metric:} A general metric is defined for heterogeneous data with missing values, which can calculate the dissimilarity between complete data and incomplete data in heterogeneous data.
		\item \textbf{Cross-type Imputation:} A new imputation paradigm is proposed, which exploits the inter-dependencies between heterogeneous attributes and takes full advantage of observed incomplete data to guide more accurate missing values inference.
		\item \textbf{Clustering with Imputation:} A clustering method for incomplete heterogeneous data sets is proposed, which establishes a complementary interaction between imputation and clustering to achieve better performance. 
	\end{itemize}

	\section{Proposed Method}\label{proposed}

We introduce a method for missing value imputation that exploits inter-dependencies among attributes and incorporates clustering results derived directly from incomplete data. The key symbols used in this paper are summarized in Table~\ref{tab:symbols}.

The method begins by clustering all objects, both complete and incomplete, using a distance matrix. Initial centroids are randomly selected and iteratively refined by replacing them with non-centroid points that minimize the replacement loss. Objects with missing values are collected into a set \( T \) and sorted in ascending order of missing values. In each iteration, the object with the fewest missing entries is selected from \( T \) as the target object. The entire dataset is clustered, and the cluster containing the target object is identified. Within this cluster, the natural neighbors \cite{NaN} of the target object are identified. Their values serve as references for imputation: the mean is used for numerical attributes, and the mode for nominal or ordinal attributes. For objects with multiple missing values, attributes exhibiting the strongest interdependence with observed attributes are imputed first \cite{zhangonline}. After each imputation, the cluster membership of the target object is updated, and the process continues with the next missing attribute using the natural neighbors from the newly assigned cluster. This iterative procedure continues until all objects in \( T \) are fully imputed. The final results comprise a complete dataset and \( k \) clusters. The overall mechanism of HMVI is illustrated in Fig.~\ref{fig:flow}.
\begin{table}[t]
\centering
\caption{Explanations of Symbols.}\label{tab:symbols}
\begin{tabular}{@{}l|l@{}}
\toprule
\textbf{Symbols} & \textbf{Explanations}      \\ \midrule
$D$                & attribute distance matrix  \\
$c_i$            & cluster centroids,$i=1,..,k$ \\
$T$ and $S$              & incomplete objects set and the whole dataset                     \\
$x_i$               &datasets objects                           \\    
$A^r$ and $o^r_m$                 &the $r$-th attribute and the $m$-th unique value of $A^r$ \\  
$n$ and $k$                 &number of objects and clusters \\
\bottomrule
\end{tabular}
\end{table}
The imputation results of HMVI are influenced by three key factors: the distance matrix between incomplete objects, the interdependence among attributes, and the natural neighbors of the target object. The distance \( \Psi(x_i, x_j) \) between two complete data objects \( x_i \) and \( x_j \) is defined as follows:
 \begin{equation}
  \Psi(x_i,x_j) = \sqrt{ {\textstyle \sum_{r=1}^{d}\Psi^r(x_i^r,x_j^r)^2} }, \label{eq:1} 
 \end{equation}
 where the $\Psi^r(x_i^r,x_j^r)$ is the dissimilarity of two possible value $x_i^r$ and $x_j^r$ of object $x_i$ and $x_j$ in attribute $A^r$. When missing values occur in $x_i$, $x_j$ or both of them, we need to modify the distance definition. This distance is calculated between the complete attribute values and then normalized to account for the missing values. Suppose $md$ is the number of attributes which are missing (in one object or the other or both), then the distance $M\Psi(x_i,x_j)$ is computed as follows:
 \begin{equation}
  M\Psi(x_i,x_j) =\sqrt[]{\sum_{r=1}^{d}\frac{d}{d-md} M\Psi^r(x^r_i,x^r_j)^2}, 
 \end{equation}
 where $M\Psi^r(x^r_i,x^r_j)$ is defined as
 \begin{equation*}
  M\Psi^r(x^r_i,x^r_j)= \begin{cases}
   0, & \text{ if $ x^r_i$ or $x^r_j$ is missing} \\
   \Psi^r(x^r_i,x^r_j) , & \text{ otherwise. } 
  \end{cases}
 \end{equation*}
 
This method assumes that distances between missing attributes equal the mean distances between complete attributes, thereby reducing the bias of conventional approaches using only available data.

The selection of attributes for imputation is prioritized based on their interdependence, which is quantified by the distance metric $\Psi^r(o_m^r,o_h^r)$, defined as:
	\begin{equation}
		\Psi^r (o^r_m,o^r_h)=\sum_{s=1}^{d}\psi ^{rs}(o^r_m,o^r_h)\cdot w^{rs}, \label{eq:2}
	\end{equation}
    where $\psi^{rs}(o^r_m , o^r_h)$ represents the dissimilarity between values $o^r_m$ and $o^r_h$ as reflected by $A^s$, which partially reflects the dependence of $A^s$ on $A^r$, and $w^{rs}$ controls the contribution of $A^s$ to $\Psi^r(o^r_m , o^r_h)$.
    If the dissimilarities between different values of $A ^r$ reflected by $A^s$ are consistently higher than those reflected by other attributes, $A^r$ and $A^s$ are considered to have stronger interdependence\cite{Graph}. The weight $w^{rs}$ is defined as:
\begin {equation}
w^{rs} = \frac {\sum_{q=1}^{K^r - 1} \sum_{c=q+1}^{K^r} \psi^{rs}(o^r_q, o^r_c)}{N^r \cdot L^r}, \label {eq:6-optimized}
\end {equation}
where \( o^r_q \) and \( o^r_c \) are the \( q \)-th and \( c \)-th unique values in the set \( O^{r<\ast>} \) (the set of non-missing unique values for attribute \( A^r \)). We denote \( K^r \) as the total number of unique values in \( O^{r<\ast>} \) (i.e., \( K^r = |O^{r<\ast>}| \)). The term \( N^r \) in the denominator of Eq. (\ref{eq:6-optimized}) represents the total number of unordered unique value pairs of \( A^r \), calculated by the combination formula based on:  
\begin {equation}
 N^r = \frac{K^r (K^r - 1)}{2}.
\end {equation}

Additionally, \( L^r \) in the denominator is the local path correction term for the pair \( (o^r_q, o^r_c) \), defined as:
\begin {equation}
 L^r = v^r_{qc} - 1.
\end {equation}

Here, \( v^r_{qc} \) refers to the number of intermediate values on the shortest path between \( o^r_q \) and \( o^r_c \), and this correction term adjusts for the topological distance between values to avoid overemphasizing indirect dependencies between attributes.
    
For a target object with multiple missing attributes, missing and complete attributes are denoted \( A^r \) and \( A^s \), respectively; the missing attribute with the strongest interdependence to any complete attribute is imputed first to maximize accuracy.  

Finally, HMVI uses natural neighbors to identify similar objects as references. Unlike KNN (which requires setting \( k \)), natural neighbors\cite{NaN} automatically determine the neighborhood and approximate \( k \) without predefined parameters. The natural neighbor of \( X_i \) is defined as:  

{\small
	\begin{equation}
		X_j \in NaN(X_i) \Leftrightarrow (X_i\in NN_r(X_j))\wedge (X_j \in NN_r(X_i)), \label{eq:NaN}
	\end{equation}}  
where \( r \) is the number of search cycles when dataset \( X \) reaches a natural steady state; \( NaN(X_i) \) is \( X_i \)’s natural neighbor set, and \( NN_r(X_i) \) is its \( r \)-nearest neighbor set. Inspired by human friendships, this method overcomes limitations of KNN\cite{knnn} and RkNN\cite{rknn}, identifying the most suitable references. Complete pseudo-code for natural neighbor search and imputation steps is in Algorithm~\ref{algorithm}.

   \begin{algorithm}[t]
    \caption{Impute missing values with HMVI}\label{algorithm}
    \KwIn{Incomplete Dataset $S$ of $N$ objects with $d$ attributes and number of clusters $k$, pure missing dataset $T$ divided from $S$}
    \KwOut{Dataset $S$ and the cluster result of $S$.}
    Sort the objects of $T$ in increasing order of missing values\; 
    \textbf{while} $T$ isn't empty \textbf{do} \\
        \quad Select an incomplete object $x_i$ from $T$\;
        \quad Compute the distance matrix $DT$ between each object in $S$ by Eq.(\ref{eq:2})\;
        \quad \textbf{Cluster} $S$ by $DT$\; 
        \quad Find the cluster $c_j$ in which the $x_i$ lies, and find natural neighbors $X_{NAN}$ of $x_i$ in $c_j$\;
        \quad \textbf{for} missing attributes of $x_i$ \textbf{do}\;
        \qquad Select the highest dependent attribute $A^r$\;
        \qquad $x^r_i$ $\gets$ imputed by mean/mode values of $A^r$ according to $x_{NaN}$\;
        \qquad Reassign $x_i$ to the nearest cluster $c_j$\;
        \quad \textbf{end for}
        \quad Delete $x_i$ in $T$, update $x_i$ in $S$.\;
    \textbf{ end while}
    \end{algorithm}

   \begin{table}[t]
	\caption{Description of the 3 datasets. $d^{<n>}$, $d^{<o>}$, $d^{<u>}$ indicate the numbers of nominal, ordinal and numerical attributes, respectively.}
	\centering
 \scalebox{0.75}{
	\begin{tabular}{c |c | c |c c c c c} 
		\toprule 
		No.&dataset       &Abbrev.& $d^{<n>}$  &$d^{<o>}$  & $d^{<u>}$  &$n$   &$k$ \\  
		\midrule 
		1 &Diagnosis     &DS      &5    &0    &1    &120  &2 \\		  
		2 &Teacher Assistant   &TA   &4    &0    &1    &151  &3 \\
		3  &Breast Cancer  &BC     &4    &2    &3    &277  &2 \\		
		\bottomrule 
	\end{tabular}  
}
	\label{tab:dataset}
\end{table}
\begin{table*}
 \centering
 \caption{Clustering performance ARI and CVI on heterogeneous datasets. Dark pink represents the best performance, light pink represents the second-best performance; The publication year of each method is indicated below the method name.}
	\resizebox{1\linewidth}{!}{    
     \label{tab:mixcluster}
\begin{tabular}{@{}ccccccccccccccccc@{}}
\toprule
\multicolumn{1}{c|}{} & \multicolumn{1}{c|}{} & \multicolumn{5}{c|}{\textbf{DS}} & \multicolumn{5}{c|}{\textbf{TA}} & \multicolumn{5}{c}{\textbf{BC}} \\ \cmidrule(l){3-17} 
\multicolumn{1}{c|}{\multirow{-2}{*}{\textbf{\begin{tabular}[c]{@{}c@{}}Missing\\ Rate\end{tabular}}}} & \multicolumn{1}{c|}{\multirow{-2}{*}{\textbf{Indicator}}} 
& \textbf{ORI} & \textbf{MMS} & \textbf{KNNMI} & \textbf{MF} & \multicolumn{1}{c|}{\textbf{HMVI}} 
& \textbf{ORI} & \textbf{MMS} & \textbf{KNNMI} & \textbf{MF} & \multicolumn{1}{c|}{\textbf{HMVI}} 
& \textbf{ORI} & \textbf{MMS} & \textbf{KNNMI} & \textbf{MF} & \textbf{HMVI} \\
\multicolumn{1}{c|}{} & \multicolumn{1}{c|}{} 
& baseline & \cite{ms},2021 & \cite{kntn},2022 & \cite{forest},2024 & \multicolumn{1}{c|}{ours} 
& baseline & \cite{ms},2021 & \cite{kntn},2022 & \cite{forest},2024 & \multicolumn{1}{c|}{ours} 
& baseline & \cite{ms},2021 & \cite{kntn},2022 & \cite{forest},2024 & ours \\ \midrule

\multicolumn{1}{c|}{} & \multicolumn{1}{c|}{ARI} 
& -0.05 & 0.131 & \cellcolor[HTML]{F5E4D9}0.186 & 0.173 & \multicolumn{1}{c|}{\cellcolor[HTML]{E7CECE}0.26} 
& 0.014 & 0.010 & 0.011 & \cellcolor[HTML]{F5E4D9}0.013 & \multicolumn{1}{c|}{\cellcolor[HTML]{E7CECE}0.0122} 
& 0.053 & \cellcolor[HTML]{E7CECE}0.043 & 0.003 & 0.040 & \cellcolor[HTML]{F5E4D9}0.014 \\ \cmidrule(lr){2-2}

\multicolumn{1}{c|}{\multirow{-2}{*}{\textbf{10\%}}} & \multicolumn{1}{c|}{CVI} 
& 0.372 & 0.360 & \cellcolor[HTML]{E7CECE}0.396 & \cellcolor[HTML]{F5E4D9}0.386 & \multicolumn{1}{c|}{0.370} 
& 0.099 & 0.102 & -0.086 & \cellcolor[HTML]{F5E4D9}0.129 & \multicolumn{1}{c|}{\cellcolor[HTML]{E7CECE}0.154} 
& 0.118 & 0.203 & \cellcolor[HTML]{E7CECE}0.210 & 0.19 & \cellcolor[HTML]{F5E4D9}0.204 \\ \midrule

\multicolumn{1}{c|}{} & \multicolumn{1}{c|}{ARI} 
& 0.022 & 0.163 & 0.171 & \cellcolor[HTML]{E7CECE}0.290 & \multicolumn{1}{c|}{\cellcolor[HTML]{F5E4D9}0.261} 
& 0.024 & 0.004 & \cellcolor[HTML]{E7CECE}0.018 & \cellcolor[HTML]{F5E4D9}0.016 & \multicolumn{1}{c|}{0.011} 
& -0.004 & \cellcolor[HTML]{E7CECE}0.043 & \cellcolor[HTML]{F5E4D9}-0.001 & -0.003 & \cellcolor[HTML]{F5E4D9}-0.001 \\ \cmidrule(lr){2-2}

\multicolumn{1}{c|}{\multirow{-2}{*}{\textbf{20\%}}} & \multicolumn{1}{c|}{CVI} 
& 0.464 & 0.347 & \cellcolor[HTML]{F5E4D9}0.359 & 0.334 & \multicolumn{1}{c|}{\cellcolor[HTML]{E7CECE}0.365} 
& 0.136 & \cellcolor[HTML]{F5E4D9}0.124 & 0.116 & 0.098 & \multicolumn{1}{c|}{\cellcolor[HTML]{E7CECE}0.181} 
& 0.19 & 0.264 & 0.232 & 0.210 & \cellcolor[HTML]{E7CECE}0.278 \\ \midrule

\multicolumn{1}{c|}{} & \multicolumn{1}{c|}{ARI} 
& 0.44 & 0.130 & \cellcolor[HTML]{E7CECE}0.329 & 0.101 & \multicolumn{1}{c|}{\cellcolor[HTML]{F5E4D9}0.272} 
& 0.024 & 0.009 & \cellcolor[HTML]{E7CECE}0.023 & \cellcolor[HTML]{F5E4D9}0.015 & \multicolumn{1}{c|}{\cellcolor[HTML]{F5E4D9}0.015} 
& 0 & \cellcolor[HTML]{E7CECE}0.058 & 0.012 & 0.007 & \cellcolor[HTML]{F5E4D9}0.014 \\ \cmidrule(lr){2-2}

\multicolumn{1}{c|}{\multirow{-2}{*}{\textbf{30\%}}} & \multicolumn{1}{c|}{CVI} 
& 0.338 & 0.331 & \cellcolor[HTML]{E7CECE}0.428 & 0.340 & \multicolumn{1}{c|}{\cellcolor[HTML]{F5E4D9}0.362} 
& 0.097 & \cellcolor[HTML]{F5E4D9}0.204 & 0.166 & 0.121 & \multicolumn{1}{c|}{\cellcolor[HTML]{E7CECE}0.206} 
& 0.226 & \cellcolor[HTML]{E7CECE}0.262 & 0.234 & \cellcolor[HTML]{F5E4D9}0.198 & 0.187 \\ \midrule

\multicolumn{1}{c|}{} & \multicolumn{1}{c|}{ARI} 
& 0.227 & 0.055 & \cellcolor[HTML]{F5E4D9}0.295 & 0.185 & \multicolumn{1}{c|}{\cellcolor[HTML]{E7CECE}0.376} 
& 0.019 & 0.003 & \cellcolor[HTML]{E7CECE}0.029 & \cellcolor[HTML]{F5E4D9}0.018 & \multicolumn{1}{c|}{0.013} 
& -0.009 & -0.003 & 0.032 & \cellcolor[HTML]{E7CECE}0.055 & \cellcolor[HTML]{F5E4D9}0.042 \\ \cmidrule(lr){2-2}

\multicolumn{1}{c|}{\multirow{-2}{*}{\textbf{40\%}}} & \multicolumn{1}{c|}{CVI} 
& 0.392 & 0.338 & \cellcolor[HTML]{E7CECE}0.468 & 0.384 & \multicolumn{1}{c|}{\cellcolor[HTML]{F5E4D9}0.438} 
& 0.113 & 0.219 & \cellcolor[HTML]{F5E4D9}0.229 & 0.194 & \multicolumn{1}{c|}{\cellcolor[HTML]{E7CECE}0.340} 
& 0.185 & \cellcolor[HTML]{E7CECE}0.296 & \cellcolor[HTML]{F5E4D9}0.249 & 0.224 & 0.238 \\ \midrule

\multicolumn{1}{c|}{} & \multicolumn{1}{c|}{ARI} 
& 0.227 & \cellcolor[HTML]{F5E4D9}0.093 & 0.077 & 0.083 & \multicolumn{1}{c|}{\cellcolor[HTML]{E7CECE}0.116} 
& 0.035 & 0.004 & \cellcolor[HTML]{E7CECE}0.015 & 0.01 & \multicolumn{1}{c|}{\cellcolor[HTML]{F5E4D9}0.009} 
& -0.002 & \cellcolor[HTML]{E7CECE}0.402 & 0.014 & \cellcolor[HTML]{F5E4D9}0.036 & -0.002 \\ \cmidrule(lr){2-2}

\multicolumn{1}{c|}{\multirow{-2}{*}{\textbf{50\%}}} & \multicolumn{1}{c|}{CVI} 
& 0.392 & 0.382 & \cellcolor[HTML]{F5E4D9}0.375 & 0.405 & \multicolumn{1}{c|}{\cellcolor[HTML]{E7CECE}0.418} 
& 0.256 & 0.168 & \cellcolor[HTML]{E7CECE}0.362 & \cellcolor[HTML]{F5E4D9}0.217 & \multicolumn{1}{c|}{0.203} 
& 0.199 & \cellcolor[HTML]{F5E4D9}0.339 & \cellcolor[HTML]{E7CECE}0.390 & 0.248 & 0.318 \\ \midrule
 
\end{tabular}
}
 \end{table*}

The time complexity of HMVI consists of two main parts: clustering and imputation. Clustering involves distance matrix computation \(O(n^2)\) and initial clustering \(O(I \cdot n \cdot k)\). Imputation includes sorting incomplete objects \(O(m \log m)\) and iterative imputation \(O(m \cdot (I \cdot n \cdot k + a \cdot n_c \log n_c))\), where \(n_c\) denotes the size of the target cluster. Overall, the total time complexity is \(O(n^2 + m \cdot I \cdot n \cdot k + m \cdot a \cdot n_c \log n_c)\), with \(n\): number of objects, \(m\): incomplete objects, \(k\): clusters, \(I\): iterations, and \(a\): missing attributes per object.
\section{Experiments}

	\begin{figure}[t]
		\centering
		\includegraphics[width=0.85\linewidth]{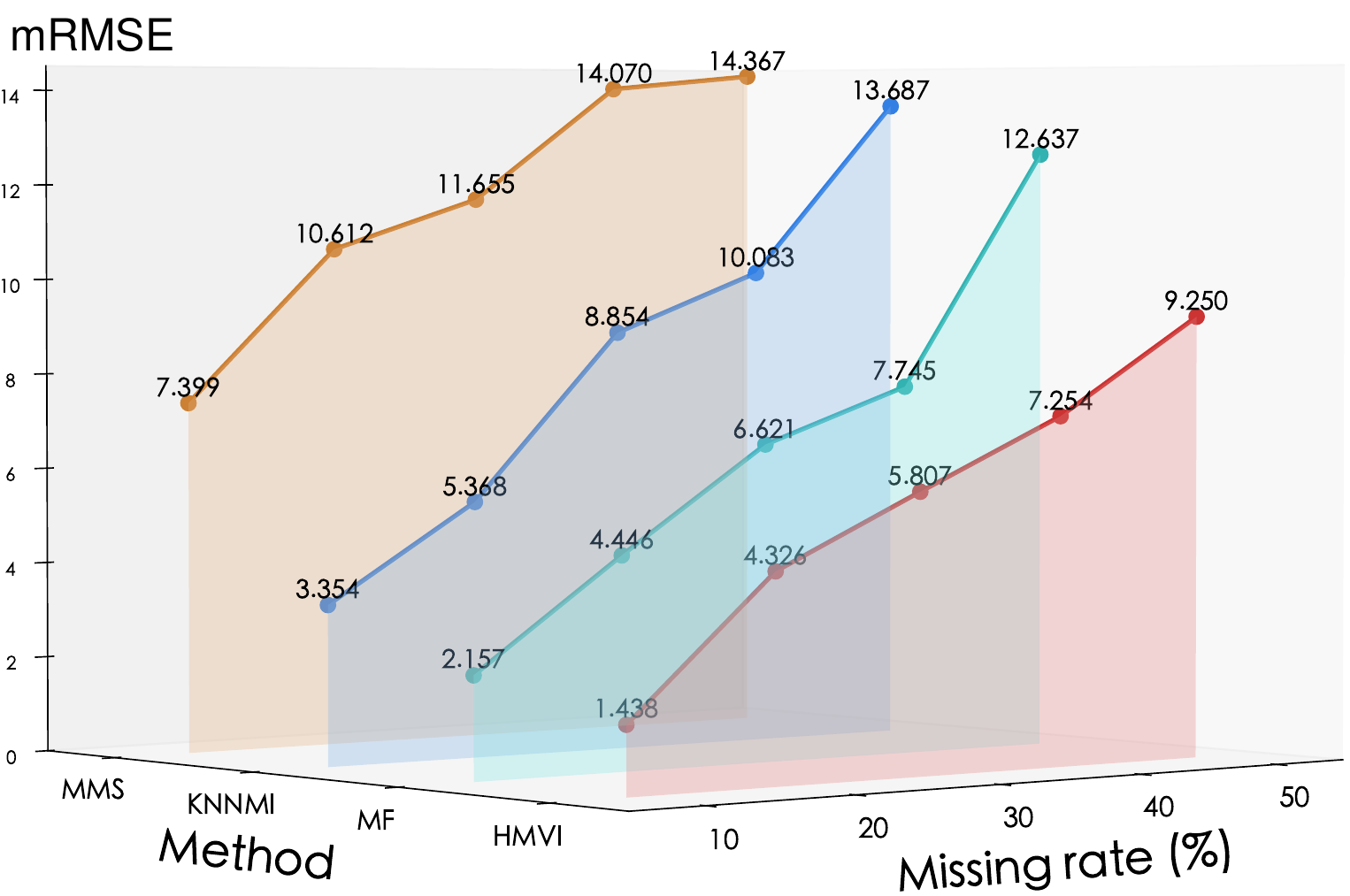}
		\caption{Evaluation of imputation accuracy for heterogeneous data (Lower value indicates better imputation performance).}
		\label{fig:cat}
	\end{figure}
 
 \textbf{Comparison Methods:} HMVI is compared with 3 well-established benchmark methods on 3 datasets with mixed-type attributes, with dataset details summarized in Table~\ref{tab:dataset}. The three benchmarks are: 1) Mean/Mode Substitution (MMS)\cite{ms}: a simple imputation approach for both numerical and categorical data; 2) MissForest (MF) \cite{forest}: a random forest regression-based method that predicts missing values using other non-missing features \cite{RFD}; 3) K-Nearest Neighbors Imputation (KNNMI)\cite{kntn}: replacing missing values with the mean of k-nearest complete neighbors. Owing to space limitations, the detailed results and discussions regarding numerical and categorical datasets are available in the supplementary material \footnote{https://github.com/gordonlok/ICASSP2026-HMVI}.

\noindent\textbf{Evaluation and Settings:} The original complete data, prior to the introduction of missing values, is denoted as the ORI. Missing values were introduced randomly at rates ranging from 10\% to 50\%. Each experimental condition was repeated 10 times, and the average results are reported. Imputation performance was evaluated using the mixed Root Mean Square Error (mRMSE). The clustering performance of K-Prototypes\cite{kprototype} on imputed data was assessed using the Adjusted Rand Index (ARI) and the Silhouette index (CVI), with clustering results on ORI serving as the reference baseline. This assessment aimed to evaluate the effectiveness of imputation methods for downstream tasks. In all clustering experiments, the number of clusters was set to the true number of classes in each dataset.

\noindent\textbf{Imputation Accuracy Evaluation:} As shown in Fig.~\ref{fig:cat}, HMVI has strong adaptability in high missing rate scenarios. By contrast, MMS always has the highest mRMSE because it only uses global statistics (mean/mode) for imputation without considering inter-data dependencies. Although MF and KNNMI utilize data correlations, they neglect the structural characteristics of heterogeneous data and fail to utilize the distribution characteristics of data for imputation.

\noindent\textbf{Clustering Performance Validity:} As shown in Table~\ref{tab:mixcluster}, HMVI performs optimally in most scenarios, fully demonstrating its imputation effectiveness and suitability for clustering tasks. Among these, for the DS dataset, as the missing rate increases, HMVI's ARI consistently maintains competitiveness, and even at a 50\% missing rate, it still showcases strong robustness against high missing rates. In the TA dataset, the ARI baseline of ORI is relatively low; yet in this scenario, HMVI can still stably match the optimal level, proving that the data imputed by HMVI can maintain the weak clustering structure of the original data.

\begin{figure}[t]
		\centering
		\includegraphics[width=1\linewidth]{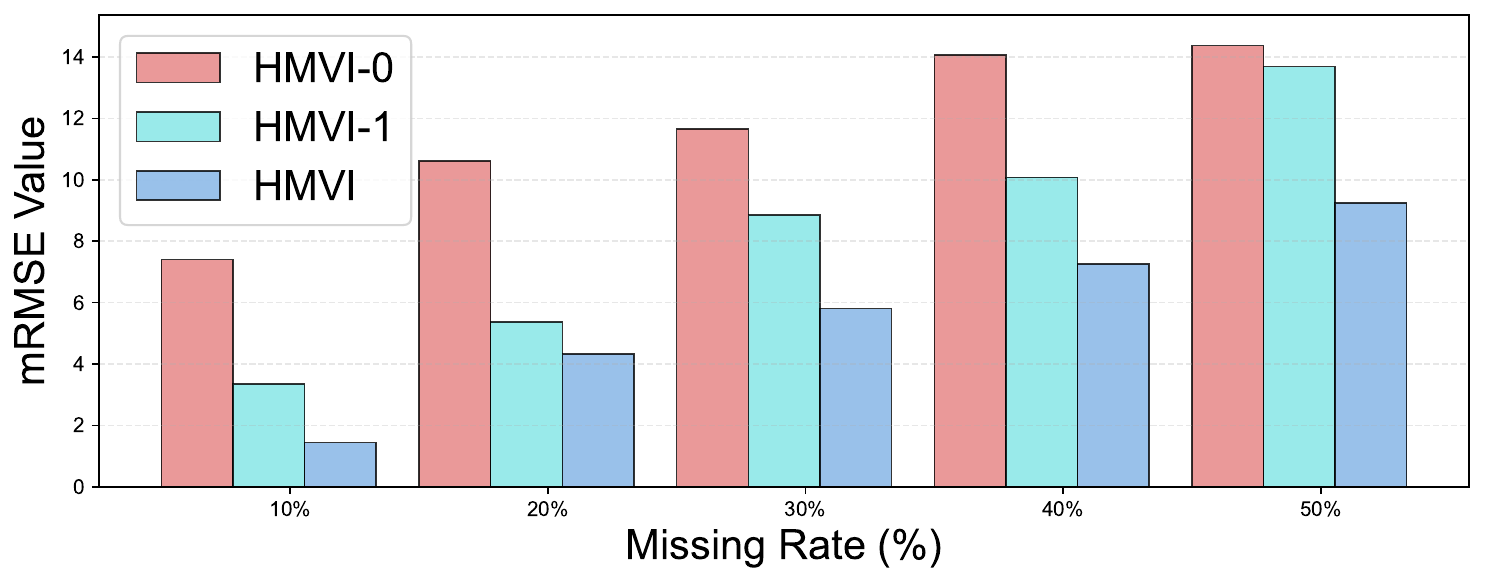}
		\caption{Evaluation of mRMSE for different ablation variants of HMVI on heterogeneous datasets.}
		\label{fig:ablation}
	\end{figure}

\noindent\textbf{Ablation Study:} Fig.~\ref{fig:ablation} evaluates the mRMSE (modified Root Mean Squared Error) of different ablation variants of HMVI on heterogeneous datasets to verify the effectiveness of the three core contributions of this paper. As the missing rate increases, the mRMSE of all variants rises, yet the full HMVI model exhibits the most stable error growth. HMVI-0 does not include neighbor search for attribute values, and HMVI-1 lacks pre-clustering; HMVI-1 outperforms HMVI-0, demonstrating the effectiveness of neighbor search, and this search relies on the foundation provided by the Unified Heterogeneous Metric. Additionally, the full HMVI model outperforms HMVI-1, which validates the value of the clustering-with-imputation method proposed in this paper, as it facilitates the complementary interaction between clustering and imputation. Specifically, pre-clustering helps neighbor search align more closely with the local data distribution, thereby enhancing imputation accuracy.
    
\section{concluding remarks}	

        The proposed HMVI method is designed to handle incomplete heterogeneous data, managing datasets containing arbitrary combinations of numerical, nominal, and ordinal attributes. It not only infers the information implied by the incomplete objects in the dataset but also effectively clusters these objects. The advantage of HMVI lies in its ability to fully exploiting remaining available information, including inter-sample and inter-attribute relationships, to infer missing values. However, we need to continue improving HMVI in the future, as it shares a common drawback with general clustering methods: the random initialization of center points can significantly impact clustering results, thereby affecting subsequent imputation results. This impact is even more profound in cases of data imbalance.
\small
\bibliographystyle{IEEEbib}
\bibliography{ICASSP2026_Paper_Templates/HMVI}

\end{document}